%% file: main.tex
\documentclass[10pt]{article}
\usepackage{robochrono-preprint}

\input{math_commands.tex}

\usepackage{booktabs,multirow,makecell,threeparttable,adjustbox,float,pifont}
\usepackage[ruled,vlined,linesnumbered]{algorithm2e}
\newcommand{\benchmarkname}{RoboChrono}
\newcommand{\cmark}{\ding{51}}
\newcommand{\xmark}{\ding{55}}
\hypersetup{pdfauthor={Yuzhou Wu; Longteng Fan; Zimeng Li; Yu Wanchan; Ting Zhang; Yiyang Ma; Shihao Li; Wei Ying; Jianbin Qin; Jiajian Jing; Fangwen Chen; Yifan Wu; Zichen Zhang; Ruiqi Yang; Weibin Kong; Yihang Xu; Haoran Liu; Zonghang He; Xuyang Liu; YiFan Xiong; Siteng Huang; Tao Xu; Zhuo Xu; Long Chen; Ruoxiang Li}}

\begin{document}
\thispagestyle{firstpage}
\begin{center}
\begin{minipage}[c]{0.31\linewidth}
  \includegraphics[height=16pt]{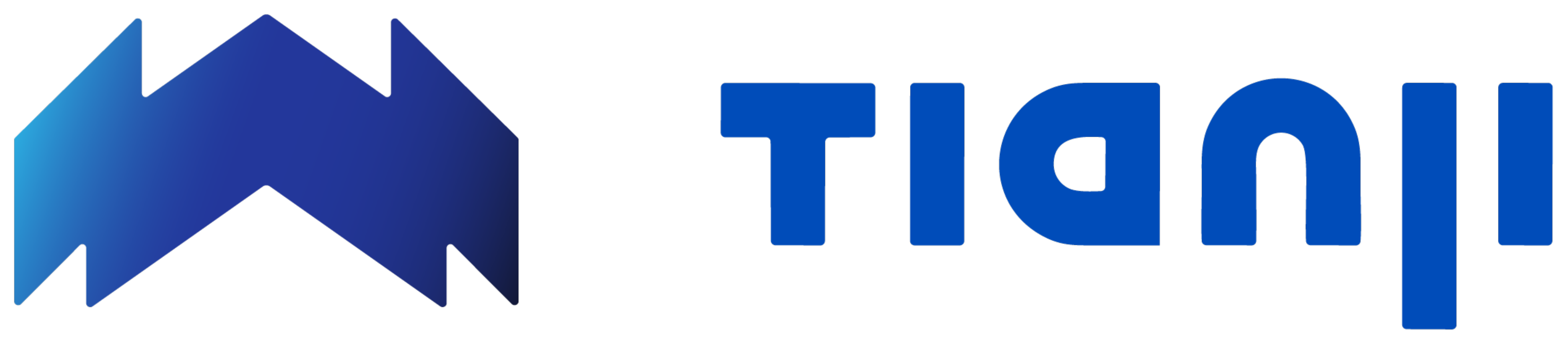}
\end{minipage}\hfill
\begin{minipage}[c]{0.28\linewidth}\centering
  \includegraphics[height=23pt]{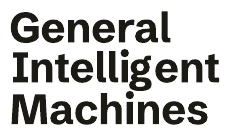}
\end{minipage}\hfill
\begin{minipage}[c]{0.34\linewidth}\raggedleft
  \includegraphics[height=23pt]{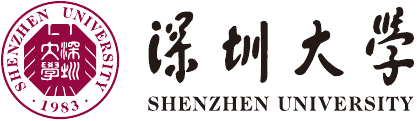}
\end{minipage}
\par\vspace{3pt}
{\color{rcgreen}\rule{\linewidth}{0.6pt}}
\par\vspace{8pt}
{\sffamily\bfseries\fontsize{22}{26}\selectfont
RoboChrono: A Real Robot Benchmark\par
for Streaming Task Understanding\par}
\vspace{12pt}
{\fontsize{10.5}{14.4}\selectfont
\person{Yuzhou Wu}{1,2,*}\authorsep
\person{Longteng Fan}{3,*}\authorsep
\person{Zimeng Li}{4,*}\authorsep
\person{Yu Wanchan}{4,*}\authorsep
\person{Ting Zhang}{2,*}\par
\person{Yiyang Ma}{5,*}\authorsep
\person{Shihao Li}{3,*}\authorsep
\person{Wei Ying}{6}\authorsep
\person{Jianbin Qin}{2}\authorsep
\person{Jiajian Jing}{4}\par
\person{Fangwen Chen}{4}\authorsep
\person{Yifan Wu}{3}\authorsep
\person{Zichen Zhang}{3}\authorsep
\person{Ruiqi Yang}{3}\authorsep
\person{Weibin Kong}{3}\par
\person{Yihang Xu}{3}\authorsep
\person{Haoran Liu}{3}\authorsep
\person{Zonghang He}{7}\authorsep
\person{Xuyang Liu}{8}\authorsep
\person{YiFan Xiong}{9}\par
\person{Siteng Huang}{10}\authorsep
\person{Tao Xu}{3}\authorsep
\person{Zhuo Xu}{3,\textdagger}\authorsep
\person{Long Chen}{3,\textdagger}\authorsep
\person{Ruoxiang Li}{2,\textdagger}\par}
\vspace{6pt}
{\fontsize{9}{11.5}\selectfont
\institution{1}{Tianji Tec.}\quad
\institution{2}{Shenzhen University}\quad
\institution{3}{General Intelligence Machine}\par
\institution{4}{Huazhong Agricultural University}\quad
\institution{5}{Yanbian University}\par
\institution{6}{South China Agricultural University}\quad
\institution{7}{Shanghai Jiao Tong University}\par
\institution{8}{Hong Kong Polytechnic University}\quad
\institution{9}{Beijing Jiaotong University}\quad
\institution{10}{Alibaba Group}\par}
\vspace{4pt}
{\small\aff{*}Equal contribution\qquad\aff{\textdagger}Corresponding authors\par}
\end{center}
\vspace{1pt}

\begin{tcolorbox}[colback=rcpale,colframe=rcpale,boxrule=0pt,arc=6pt,
  left=13pt,right=13pt,top=10pt,bottom=10pt,before skip=0pt,after skip=13pt]
{\sffamily\bfseries\color{rcink}Abstract}\par\smallskip
\begingroup\setlength{\parindent}{0pt}
\input{Sections/0_Abstract}
\endgroup
\vspace{7pt}
\noindent\begin{minipage}[c]{0.86\linewidth}
  \fontsize{8.7}{12}\selectfont\raggedright
  \textbf{Website:} \href{https://continuity3.github.io/robochrono/}{continuity3.github.io/robochrono}\par
  \textbf{GitHub:} \href{https://github.com/mfan-res/ROBOCHRONO}{github.com/mfan-res/ROBOCHRONO}\par
  \textbf{Tianji dataset:} \href{https://huggingface.co/datasets/gimai/RC-Tianji}{huggingface.co/datasets/gimai/RC-Tianji}\par
  \textbf{GIM dataset:} \href{https://huggingface.co/datasets/gimai/RC-GIM}{huggingface.co/datasets/gimai/RC-GIM}
\end{minipage}\hfill
\begin{minipage}[c]{0.12\linewidth}\centering
  \includegraphics[width=\linewidth]{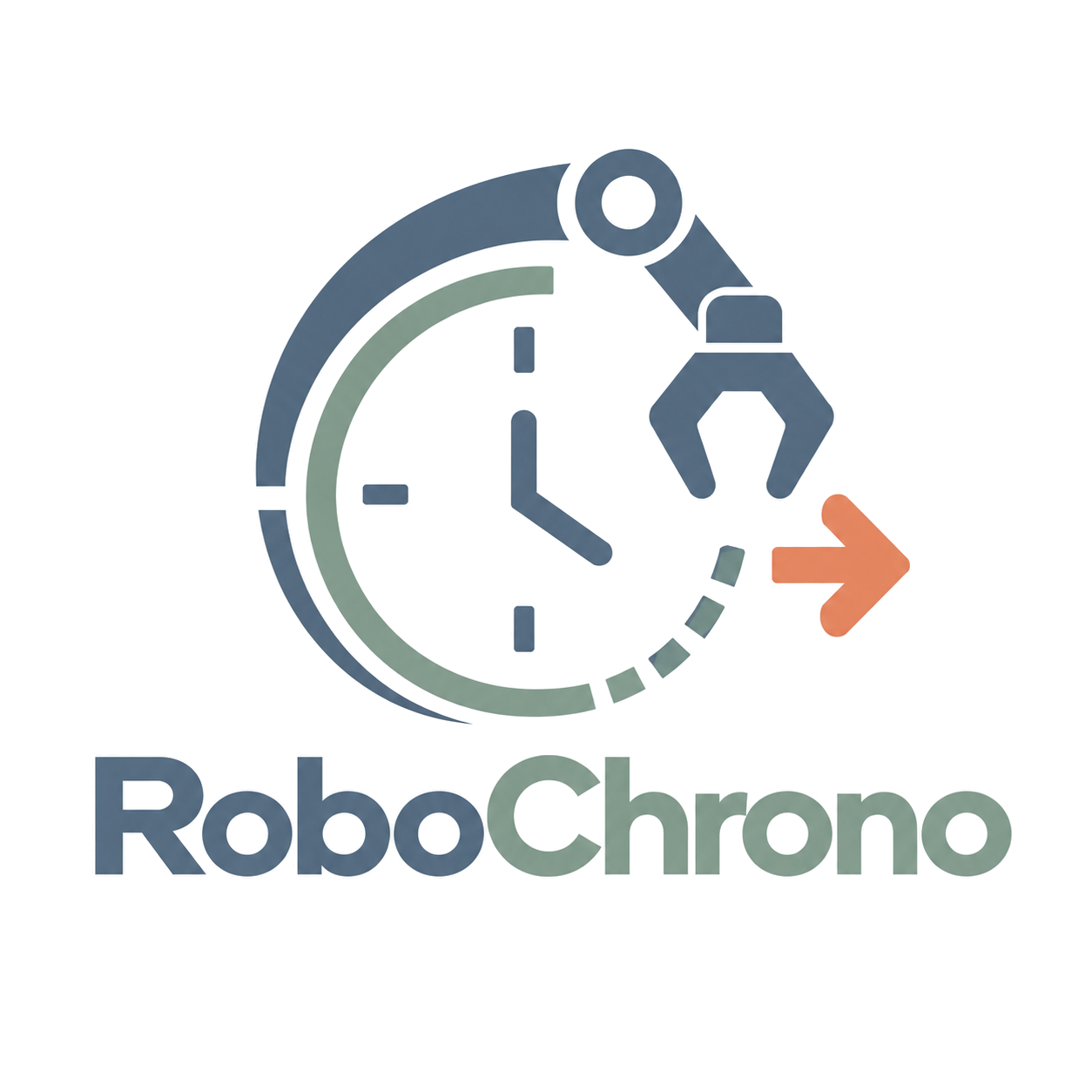}
\end{minipage}
\end{tcolorbox}

\input{Sections/1_Introduction}

\input{Sections/2_Related_Work}
\input{Sections/3_Methodology}
\input{Sections/4_Experiments}
\input{Sections/5_Conclusion}

\section*{Resources}
The project website is \url{https://continuity3.github.io/robochrono/}.
The code repository is \url{https://github.com/mfan-res/ROBOCHRONO}.
The Tianji and GIM dataset repositories are
\url{https://huggingface.co/datasets/gimai/RC-Tianji} and
\url{https://huggingface.co/datasets/gimai/RC-GIM}, respectively.

\section*{AI Use Statement}
Generative AI tools were used only for language editing and writing assistance.
They were not used for data generation, experimental execution, result analysis,
or scientific decision making. All AI-assisted content was reviewed by the
authors, who take full responsibility for the final manuscript.

\bibliographystyle{unsrtnat}
\bibliography{references}
\clearpage
\appendix
\section{Supplementary Material}
\input{Sections/6_appendix}

\end{document}

%% file: math_commands.tex
\usepackage{amsmath,amsfonts,bm}

\def\eqref#1{equation~\ref{#1}}

\def\1{\bm{1}}

\DeclareMathAlphabet{\mathsfit}{\encodingdefault}{\sfdefault}{m}{sl}
\SetMathAlphabet{\mathsfit}{bold}{\encodingdefault}{\sfdefault}{bx}{n}



%% file: Sections/0_Abstract.tex
Understanding ongoing robot manipulation requires models to interpret visual observations in relation to interaction history and task progress. We introduce RoboChrono, a benchmark for streaming task understanding comprising 39 scenarios and 34,713 evaluation instances, constructed from real robot executions and complementary bare-hand human recordings.
The benchmark evaluates seven tasks grouped into recognition,
alignment, and temporal grounding, covering action understanding and anticipation, visual correspondence, temporal ordering, and action localization. Zero-shot evaluation of 18 vision-language models reveals substantial differences across tasks. GPT-6-Astra achieves 98.3\% accuracy on Frame Matching but 68.3\%
on Frame Ordering, while RynnBrain1.1-122B-A10B exhibits a larger
gap, reaching 95.4\% and 32.9\%, respectively.
Input ablations on matched questions with five open-weight models
further reveal distinct dependencies on visual evidence:
removing visual observations reduces Current Action Recognition
accuracy by 22.1 percentage points, whereas Next Action Prediction
decreases by only 0.7 points.
These findings show that strong visual matching does not consistently
coincide with strong temporal ordering, and suggest that next-action
prediction can be supported by task and action priors even when
visual evidence is unavailable.
RoboChrono provides a diagnostic setting for examining these differences,
highlighting the need for capability-specific evaluation beyond
aggregate scores when assessing task understanding in robot manipulation. 

%% file: Sections/1_Introduction.tex
\section{Introduction}

Understanding an ongoing robot manipulation task requires more than
recognizing what is visible in the current frame. An embodied agent must reason over the preceding interaction history to infer the current task state, distinguish completed from ongoing actions, and anticipate what may happen
next. Prior work on egocentric video has similarly shown the importance of
temporal context for fine-grained action understanding and anticipation%
~\citep{damen2018scaling, furnari2019would, jia2022egotaskqa}.
The problem is particularly challenging from robot-mounted viewpoints, where
occlusion, motion, and partial observability can limit the evidence available
from any single observation or camera%
~\citep{jangir2022look, khazatsky2024droid}.
We study streaming temporal task understanding: interpreting the progress of robot manipulation using only visual observations available up to the queried moment, without access to any subsequent frames or evidence. \par

Existing benchmarks have advanced several complementary aspects of embodied visual reasoning. Human egocentric benchmarks evaluate temporal and causal reasoning over recorded activities~\citep{jia2022egotaskqa}, while video based spatial benchmarks focus on understanding spatial relationships from visual observations~\citep{yang2025vsi}. Robot centric datasets and benchmarks further connect visual understanding with long horizon reasoning, planning, and physical interaction~\citep{sermanet2023robovqa,chen2025robo2vlm,bao2026actioneqa}, and recent embodied foundation model benchmarks broaden evaluation toward multimodal reasoning in physical environments~\citep{ji2025robobrain,rynnbrain2026}. However, these settings largely emphasize reasoning over completed or retrospectively available observations, where later frames may provide evidence for interpreting earlier events. In an ongoing robot execution, such future information is unavailable, and models must infer the current task state, distinguish completed from ongoing actions, recover temporal structure, and anticipate subsequent behavior from the observation history available at the queried moment. This creates a distinct evaluation challenge that is not fully captured by conventional video question answering or full video reasoning. We therefore focus on streaming temporal task understanding and ask whether multimodal models can reliably interpret the temporal progress of real robot manipulation under explicitly restricted observation histories. Table~\ref{tab:robot_ego_benchmark_comparison} compares RoboChrono with representative egocentric and robot-centric datasets and benchmarks across data sources and annotation types.\par

\begin{figure}[t]
    \centering
    \includegraphics[width=\textwidth]
    {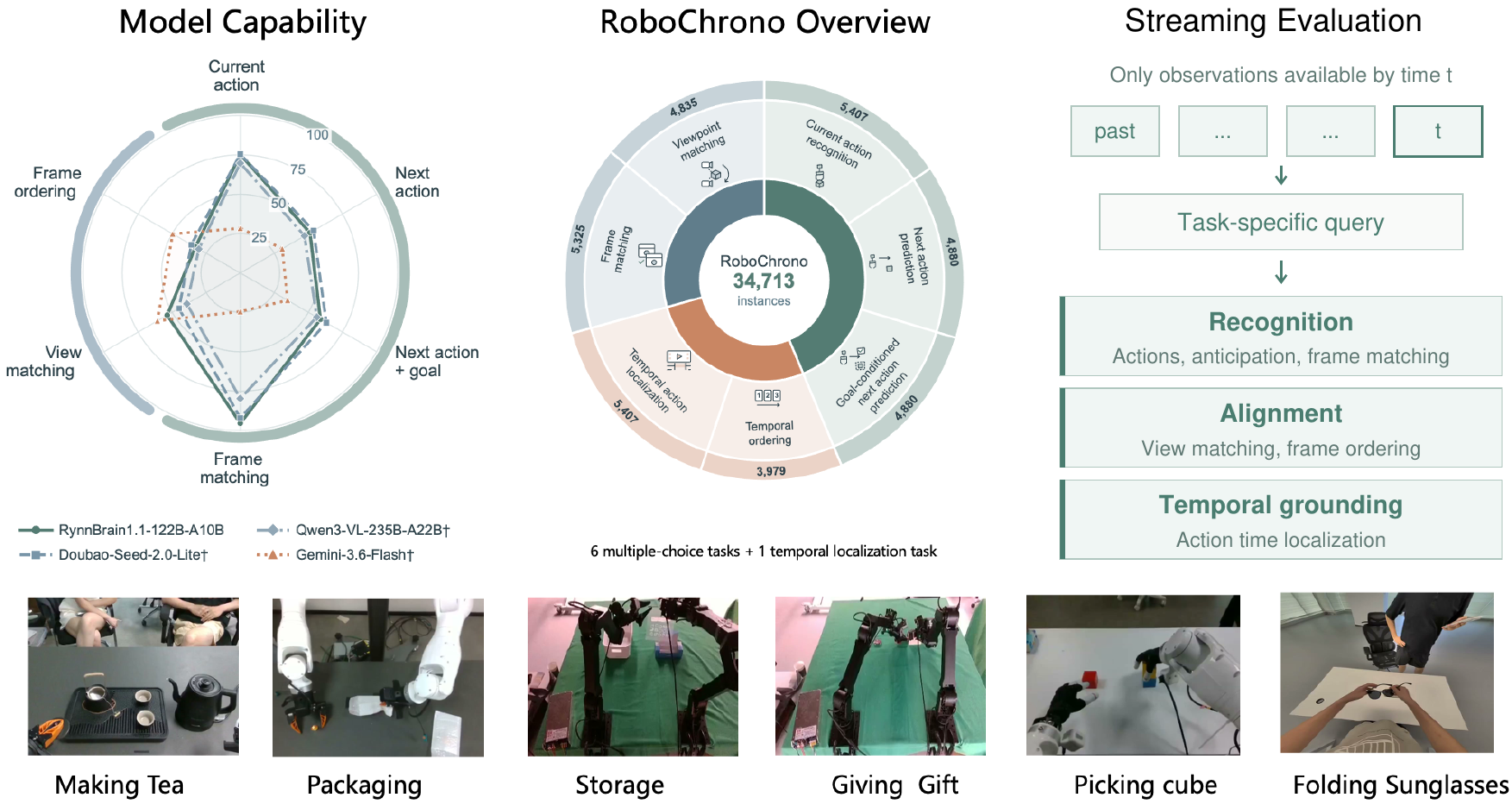}
    \caption[Overview of RoboChrono.]{
        Overview of RoboChrono.
        Left: Performance profiles of four representative
        multimodal models across the six multiple-choice evaluation
        tasks.
        Center: Distribution of 34,713 evaluation instances
        across seven tasks.
        Right: Streaming evaluation under a causal observation boundary;
        only observations available up to the queried moment are provided.
    }
    \label{fig:teaser}
\end{figure}

We introduce \textbf{RoboChrono}, a real-world robot benchmark for
streaming temporal task understanding. RoboChrono contains 39 manipulation
scenarios, 1,816 successful execution videos, approximately 25 hours of
real-world manipulation data, and 34,713 evaluation instances. As illustrated
in Figure~\ref{fig:teaser}, the benchmark evaluates seven tasks
covering action interpretation and anticipation, visual evidence association,
and temporal organization. In the streaming setting, models can only access
observations available up to the queried moment and cannot use future visual
evidence. We evaluate 18 open-source and commercial multimodal models across
all 39 scenarios under a zero-shot protocol. Our results reveal substantial differences in capability profiles,
particularly on temporal ordering, localization, and future action reasoning,
which aggregate scores often obscure.\par

Our contributions are threefold. First, we introduce \textbf{RoboChrono}, a
large-scale real robot benchmark specifically designed for streaming temporal
task understanding. Second, we establish a unified evaluation framework with
seven complementary tasks that assess action understanding, anticipation,
cross-view and temporal association, ordering, and temporal localization under
controlled observation histories. Third, we provide a systematic zero-shot
evaluation of 18 multimodal models across all 39 scenarios, revealing
persistent limitations in understanding the temporal structure and progression
of real-world robot manipulation, particularly in ordering and localization.\par

\input{Tables/first}

%% file: Tables/first.tex
\providecommand{\cmark}{\textcolor{green!55!black}{\checkmark}}
\providecommand{\xmark}{\textcolor{red!75!black}{$\times$}}
\providecommand{\pmark}{\textcolor{orange!85!black}{$\triangle$}}
\providecommand{\tbd}{\textcolor{gray}{--}}

\begin{table}[t]
\centering
\caption[Comparison of egocentric and embodied robot task benchmarks and datasets.]{
Comparison of egocentric and embodied robot task benchmarks and datasets.
We compare data sources, robot-mounted views, task collection and annotation
protocols, and evaluation scale. RoboChrono uses real-robot executions with
complementary onboard views and human-defined temporal annotations, evaluated
under restricted observation histories.
}
\label{tab:robot_ego_benchmark_comparison}

\vspace{1mm}

\begingroup
\scriptsize
\setlength{\tabcolsep}{2.4pt}
\renewcommand{\arraystretch}{1.02}

\resizebox{\textwidth}{!}{%
\begin{tabular}{@{}llccccccccl@{}}
    \toprule
    \textbf{Benchmark / Dataset}
    & \textbf{Primary focus}
    & \textbf{Data source}
    & \multicolumn{2}{c}{\textbf{Robot-mounted view}}
    & \multicolumn{3}{c}{\textbf{Task collection}}
    & \multicolumn{2}{c}{\textbf{Curation}}
    & \textbf{Evaluation scale} \\
    \cmidrule(lr){3-3}
    \cmidrule(lr){4-5}
    \cmidrule(lr){6-8}
    \cmidrule(lr){9-10}

    &
    & \shortstack{\textbf{Real}\\\textbf{robot}}
    & \textbf{Head}
    & \textbf{Wrist}
    & \shortstack{\textbf{All}\\\textbf{teleop}}
    & \shortstack{\textbf{Task}\\\textbf{exec.}}
    & \shortstack{\textbf{Task}\\\textbf{QC}}
    & \shortstack{\textbf{Human}\\\textbf{task}}
    & \shortstack{\textbf{Manual}\\\textbf{labels}}
    & \\
    \midrule

    \rowcolor{blue!7}
    \multicolumn{11}{c}{%
        \textit{(i) Human-worn or scene-centric embodied cognition benchmarks}%
    } \\

    EgoTaskQA~\cite{jia2022egotaskqa}
    & Human task reasoning
    & \xmark & \xmark & \xmark
    & \xmark & \cmark & \pmark
    & \cmark & \pmark
    & 2k videos / 40k QA \\

    VSI-Bench~\cite{yang2025vsi}
    & Spatial cognition
    & \xmark & \xmark & \xmark
    & \xmark & \xmark & \xmark
    & \pmark & \pmark
    & 288 videos / 5k+ QA \\

    RoboSpatial~\cite{song2025robospatial}
    & Spatial / affordance QA
    & \xmark & \xmark & \xmark
    & \xmark & \xmark & \xmark
    & \xmark & \xmark
    & 1M images / 3M QA \\

    ERQA~\cite{deepmind2025erqa}
    & Embodied reasoning QA
    & \pmark & \pmark & \pmark
    & \xmark & \xmark & \xmark
    & \cmark & \cmark
    & 400 QA \\

    MMSI-Bench~\cite{yang2026mmsi}
    & Multi-image spatial QA
    & \pmark & \pmark & \pmark
    & \xmark & \xmark & \xmark
    & \cmark & \cmark
    & 1,000 QA \\

    RynnBrain-Bench~\cite{rynnbrain2026}
    & Cognition + localization
    & \xmark & \xmark & \xmark
    & \xmark & \xmark & \xmark
    & \pmark & \pmark
    & 3,616 clips / 12k QA \\

    \midrule

    \rowcolor{green!7}
    \multicolumn{11}{c}{%
        \textit{(ii) Physical-robot embodied benchmarks and demonstration resources}%
    } \\

    JRDB~\cite{martinmartin2021jrdb}
    & Perception / tracking
    & \cmark & \cmark & \xmark
    & \xmark & \xmark & \xmark
    & \xmark & \cmark
    & 64 min / 3.5k tracks \\

    RoboVQA~\cite{sermanet2023robovqa}
    & QA + long-horizon planning
    & \cmark & \cmark & \xmark
    & \pmark & \cmark & \pmark
    & \cmark & \pmark
    & 238 h / 829.5k pairs \\

    DROID$^{\dagger}$~\cite{khazatsky2024droid}
    & Robot demonstrations
    & \cmark & \xmark & \cmark
    & \cmark & \cmark & \cmark
    & \cmark & \cmark
    & 76k traj. / 350 h / 86 tasks \\

    Open-X VQA / Robo2VLM-1~\cite{chen2025robo2vlm}
    & Robot interaction QA
    & \cmark & \pmark & \pmark
    & \cmark & \cmark & \pmark
    & \pmark & \xmark
    & 176k traj. / 684.7k QA \\

    ShareRobot (Aff./Traj.)~\cite{ji2025robobrain}
    & Planning / affordance / trajectory
    & \cmark & \pmark & \pmark
    & \pmark & \cmark & \cmark
    & \pmark & \pmark
    & 51.4k inst. / 1.028M QA \\

    ActionEQA~\cite{bao2026actioneqa}
    & Action-state reasoning
    & \cmark & \pmark & \cmark
    & \cmark & \cmark & \pmark
    & \pmark & \pmark
    & 2,629 eps. / 8,795 QA \\

    \midrule

    \rowcolor{yellow!15}
    \textbf{RoboChrono (Ours)}
    & \textbf{Robot-ego temporal understanding}
    & \cmark & \cmark & \cmark
    & \pmark & \cmark & \cmark
    & \cmark & \cmark
    & \textbf{39 scenarios / 34,713 QA} \\

    \bottomrule
\end{tabular}%
}

\endgroup

\vspace{1mm}

\parbox{\textwidth}{%
    \footnotesize
    \textbf{Definitions.}
    \cmark: yes; \xmark: no; \pmark: partial or mixed.
    ``Real robot'' denotes physical robot data; ``Head'' and ``Wrist'' denote
    global and end-effector-mounted views, respectively.
    ``All teleop'' and ``Manual labels'' indicate fully human-teleoperated
    executions and human-produced annotations, respectively.
    ``Task QC'' denotes execution quality control. RoboChrono has mixed
    collection: robot executions are teleoperated, while five scenarios
    contain bare-hand human demonstrations.
}

\end{table}

%% file: Sections/2_Related_Work.tex
\section{Related Work}

\noindent\textbf{Egocentric Video Understanding.}
Egocentric video has become an important setting for studying temporal understanding from the first person perspective. Large scale datasets such as Ego4D~\citep{grauman2022ego4d} and Ego Exo4D~\citep{grauman2024egoexo4d} support research on perception, episodic memory, action recognition, and forecasting. Recent benchmarks further emphasize higher level task reasoning. EgoTaskQA~\citep{jia2022egotaskqa} evaluates temporal and causal reasoning over actions and state changes, while EgoSchema~\citep{mangalam2023egoschema}, EgoPlan Bench~\citep{chen2023egoplan}, and VidEgoThink~\citep{cheng2024videgothink} study long-video question answering, task planning, and embodied reasoning. However, these benchmarks are primarily built from human-worn videos, whose viewpoints and motion patterns differ from those of cameras mounted on physical robots.\par

\noindent\textbf{Robot-Centric Embodied Understanding.}
Recent work has extended visual reasoning from human activities to robot interaction data. RoboVQA~\citep{sermanet2023robovqa} constructs large-scale video question answering data from both human and robot embodiments to study long-horizon robotic reasoning. Robo2VLM~\citep{chen2025robo2vlm} converts real robot trajectories into visual question answering samples by extracting manipulation phases and supervision from robot states, end-effector poses, gripper measurements, and force signals. ActionEQA~\citep{bao2026actioneqa} evaluates the relationship between semantic actions and physical state transitions. More recent embodied benchmarks, including RoboBench~\citep{luo2025robobench}, RoboBrain~\citep{ji2025robobrain}, and RynnBrain~\citep{rynnbrain2026}, further broaden evaluation toward embodied perception, spatial reasoning, affordance understanding, and planning. These efforts demonstrate the importance of evaluating multimodal models on physically grounded robot data, although their data sources, embodiments, camera configurations, and annotation pipelines are often heterogeneous or constructed for broad embodied capabilities rather than controlled temporal task understanding in streaming settings.\par

\noindent\textbf{Streaming Video Understanding.}
Most video understanding benchmarks assume that the complete video is available before answering a query. This offline setting differs from embodied agents that must interpret observations while an interaction is still unfolding. StreamingBench~\citep{lin2024streamingbench} evaluates multimodal models with questions presented at different timestamps, targeting real-time perception and accumulated context. OVBench~\citep{huang2025onlinevideo} further organizes online video understanding around perception, memory, and prediction over past, present, and future information. Recent work also reveals a tradeoff between immediate perception and long-range memory~\citep{shen2026simple}. However, existing streaming benchmarks mainly use general videos and human-centered activities rather than physical robot manipulation with explicit task structure.\par

Different from existing egocentric, robotic, and streaming benchmarks, RoboChrono focuses on streaming task understanding from controlled real robot executions. Robot episodes are collected under a unified teleoperation and quality control protocol with complementary head and wrist observations; the dataset also includes five bare-hand human scenarios. The benchmark covers both gripper and dexterous-hand manipulation and provides human-annotated temporal events for every execution. These annotations support evaluation of task-stage understanding, event localization, question answering, and next-action prediction at meaningful moments during execution. To our knowledge, RoboChrono is the first benchmark to jointly evaluate these properties in a unified real-robot setting under restricted observation histories.\par

%% file: Sections/3_Methodology.tex
\section{Benchmark Design and Evaluation}

\subsection{Benchmark Overview}

\begin{figure}[t]
    \centering
    \includegraphics[width=\textwidth]{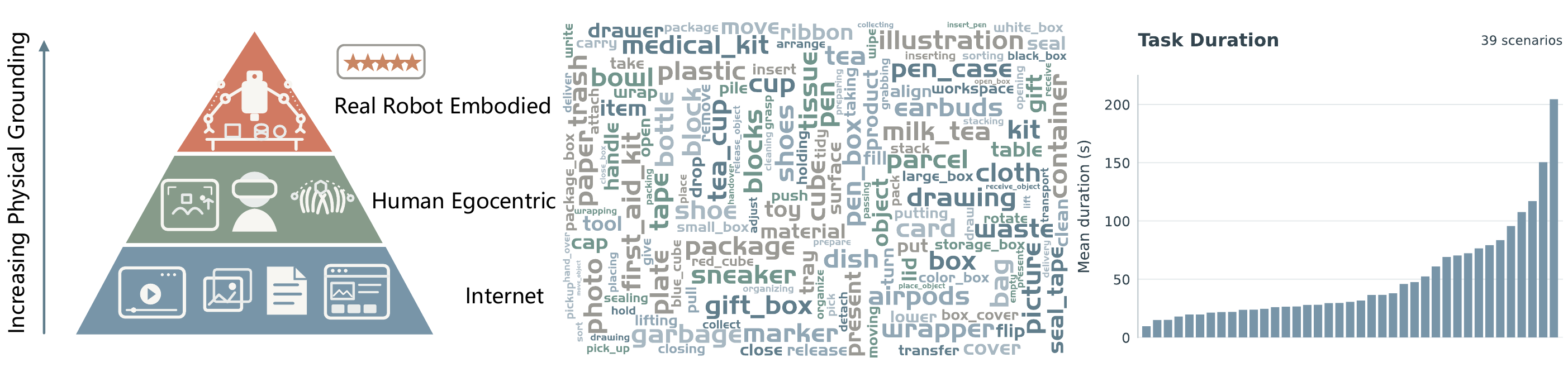}
    \caption[Overview of the task coverage in \benchmarkname{}.]{
        Overview of the task coverage in \benchmarkname{}.
        The benchmark includes diverse manipulation objects and action
        primitives drawn from 39 real-world manipulation scenarios.
    }
    \label{fig:task_coverage}
\end{figure}

\begin{figure}[t]
\centering
\includegraphics[width=\textwidth]{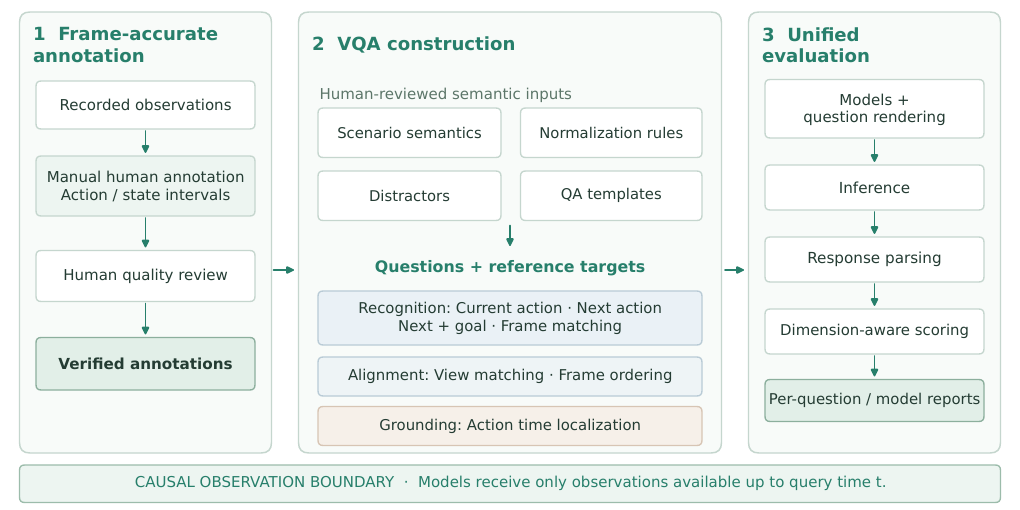}
\caption{
Overview of the \benchmarkname{} construction and evaluation pipeline.
Human-annotated action intervals support task-specific question construction
and unified model evaluation under a causal observation boundary.
}
\label{fig:benchmark_pipeline}
\end{figure}

\benchmarkname{} evaluates whether multimodal models can understand the
progress of real-world robot manipulation from progressively revealed
visual observations. The benchmark focuses on three capabilities: understanding ongoing actions and anticipating subsequent
behavior, establishing correspondences across observations and camera
views, and localizing task-relevant events in time. At an evaluation time $t$, a model receives the observed video prefix
$\mathcal{O}_{\leq t}$ and a task-specific query $q_t$. It must answer
using the available observations without accessing future frames.
Depending on the evaluation task, the required output is either a
multiple-choice answer or a temporal interval. Explicit task goals are provided for goal-conditioned next-action prediction. The benchmark comprises seven evaluation tasks constructed from
temporally annotated manipulation executions.
Figure~\ref{fig:task_coverage} illustrates the range of manipulation
objects and action primitives covered by \benchmarkname{}, while
Figure~\ref{fig:benchmark_pipeline} summarizes the complete benchmark
construction and evaluation pipeline, from frame-accurate temporal
annotation to VQA generation and unified zero-shot model evaluation.\par

\subsection{Data Collection and Temporal Annotation}

We design 39 manipulation tasks, including 29 gripper based tasks,
5 dexterous hand tasks, and 5 ego hand tasks. The task set spans
object grasping, picking, placing, transferring, stacking, packaging,
cleaning, and tool use, covering both object-level rearrangement and fine-grained hand-object interaction. The objects include everyday
items, containers, tools, and packaged goods.\par

Robot task executions are collected through human teleoperation under a
shared collection protocol, except for the five barehand scenarios, which are recorded from a person performing the task bare-handed. The dataset, as part of this benchmark, contains 1,816 successful execution videos and approximately 25 hours of real manipulation data. Repeated executions introduce variation in object configurations,
motion trajectories, and interaction patterns within each task in a consistent manner.

Visual observations are recorded from robot mounted or near egocentric
cameras. Multi view recordings provide complementary head and wrist
views, capturing workspace context and local manipulation details,
respectively. Single view recordings are also retained. When multiple
camera streams are available, they are synchronized within each
execution. All recorded episodes are manually inspected after
collection before inclusion in the benchmark and associated with task level metadata.

Each execution is annotated with task-relevant action and
state intervals. We represent the annotations as
$\mathcal{A} = \{(s_i, e_i, z_i)\}_{i=1}^{K}$, where $s_i$ and $e_i$
denote the start and end times of an interval, and $z_i$ describes the action or task state. Examples include approaching an
object, establishing a grasp, moving an object toward a target,
completing placement, and reaching a task-specific goal state.
We use temporal intervals rather than timestamps to preserve
the duration of actions and persistent task states. These annotations
provide the reference semantics and temporal boundaries used to sample
evaluation points and construct task-specific questions and answers.
The benchmark contains 34,713 evaluation instances spanning seven
evaluation tasks across 39 manipulation scenarios.\par

\subsection{Evaluation Protocol}
\label{sec:task_eval_protocol}

We evaluate all models under a causal observation setting that reflects
the information available to an embodied agent during online interaction.
For an evaluation instance at time $t$, the model can only access visual
observations available at or before $t$. Future observations are never
included in the model input. For next action prediction, later annotations
may be used during benchmark construction to determine the reference answer,
but the corresponding future visual evidence remains hidden at inference
time. Matching, ordering, and temporal localization instances are similarly
constructed using only observations available within the permitted
observation range. \par

The seven tasks instantiate this causal protocol with different forms of
visual evidence and reference targets. Table~\ref{tab:task_inputs}
summarizes the model inputs and target outputs for each task. \par

\begin{table}[htbp]
\centering
\footnotesize
\setlength{\tabcolsep}{3.5pt}
\renewcommand{\arraystretch}{1.12}
\caption{Task inputs and reference targets in \benchmarkname{}.}
\label{tab:task_inputs}
\begin{tabular}{p{0.30\linewidth} p{0.37\linewidth} p{0.28\linewidth}}
\hline
Task & Model input & Reference target \\
\hline
Current action &
Video prefix $\mathcal{O}_{\leq t}$ and answer choices &
Action or state at time $t$ \\

Next action &
Video prefix $\mathcal{O}_{\leq t}$ and answer choices &
Next action executed after $t$ \\

Goal-conditioned next action &
Video prefix $\mathcal{O}_{\leq t}$, task goal, and answer choices &
Next executed action under the goal \\

Frame matching &
Query state or frame and candidate frames from $\mathcal{O}_{\leq t}$ &
Frame matching the queried state \\

View matching &
Query image and synchronized candidates from other views &
Image from another view at the same time step \\

Frame ordering &
Candidate frames from $\mathcal{O}_{\leq t}$ &
Correct temporal order \\

Action time localization &
Video observation, queried action/state, and video duration &
Interval $[s_i,e_i]$ in seconds \\
\hline
\end{tabular}
\end{table}

View matching is generated only when synchronized multi-view recordings
are available; single-view recordings are excluded from this task rather
than counted as incorrect. When an action label appears multiple times in
one execution, the queried instance is disambiguated by its temporal
position and annotation index, so that each question refers to a unique
annotated interval.\par

All models are evaluated in a strictly zero-shot setting without
benchmark-specific supervised fine-tuning or adaptation. We use the same
task definitions, answer spaces, and scoring rules across models.
Prompt templates are kept consistent whenever permitted by the corresponding
inference interfaces. Closed-source models are evaluated through their
officially available interfaces, while open-source models are evaluated
locally using their publicly released checkpoints.\par

For the six multiple-choice tasks, performance is measured using accuracy.
Given $N$ valid evaluation instances, the accuracy is computed as

\begin{equation}
    \mathrm{Acc}
    =
    \frac{1}{N}
    \sum_{i=1}^{N}
    \mathbf{1}
    \left[
        \hat{y}_i = y_i^*
    \right],
\end{equation}

where $\hat{y}_i$ is the model prediction and $y_i^*$ is the annotated
reference answer. Task--recording combinations for which a question is not
applicable, such as view matching for single-view recordings, are excluded
from the corresponding evaluation set rather than treated as incorrect
predictions. \par

For action time localization, the model predicts a temporal interval
$\hat{I}=[\hat{s},\hat{e}]$ corresponding to the queried action or task
state. The localization prompt specifies the total duration of the video
and requires the predicted interval to be expressed in seconds. We measure
the overlap between the predicted interval $\hat{I}$ and the annotated
reference interval $I^*$ using temporal Intersection over Union:

\begin{equation}
    \mathrm{tIoU}(\hat{I}, I^*)
    =
    \frac{|\hat{I} \cap I^*|}
         {|\hat{I} \cup I^*|},
\end{equation}

where $|\cdot|$ denotes temporal duration. Since each query requires a
single predicted interval, we report Recall@1 at a tIoU threshold of
$\tau=0.5$:

\begin{equation}
    R@1(\tau)
    =
    \frac{1}{N_{\mathrm{loc}}}
    \sum_{i=1}^{N_{\mathrm{loc}}}
    \mathbf{1}
    \left[
        \mathrm{tIoU}(\hat{I}_i, I_i^*) \geq \tau
    \right],
\end{equation}

where $N_{\mathrm{loc}}$ denotes the number of valid localization queries.
We refer to this metric as \mbox{tIoU@0.5} throughout the paper. Finally, we distinguish invalid evaluation runs from valid but low-scoring
predictions. A \mbox{model--task} run with a response parse failure rate of
50\% or higher is treated as an unsuccessful run and is excluded from
reported comparisons. Valid runs are retained regardless of their resulting
score, so that poor task performance is not conflated with response parsing
failure.\par

%% file: Sections/4_Experiments.tex
\section{Experiments and Results}
\label{sec:experiments}

\subsection{Experimental Setup}
\label{sec:experimental_setup}

We evaluate 18 multimodal models on \benchmarkname{} under the
zero-shot protocol described in
Section~\ref{sec:task_eval_protocol}.
All models use the same task definitions and scoring rules,
without benchmark-specific supervised fine-tuning.
We report accuracy for the six multiple-choice tasks and
Recall@1 at a tIoU threshold of 0.5 for Action Time Localization.
The multiple-choice aggregate is computed over the applicable
questions and does not include Action Time Localization,
which is scored separately using this localization metric.\par

\subsection{Main Results}
\label{sec:main_results}

\input{Tables/Main_exp}

\begin{figure}[t]
    \centering
    \includegraphics[width=\linewidth]{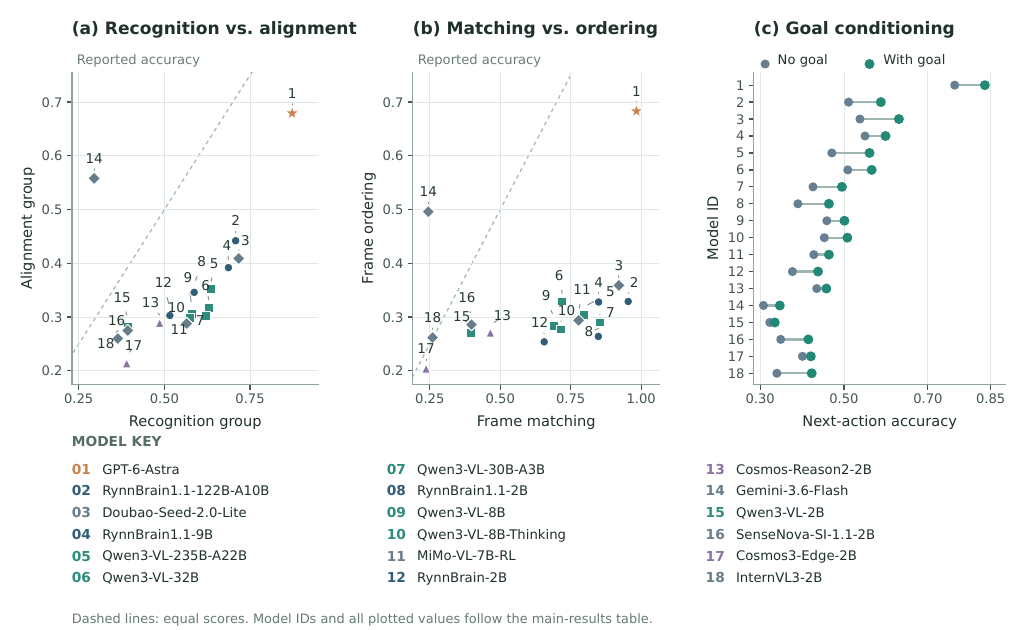}
    \caption{
        Diagnostic analysis of model capabilities on \benchmarkname{}.
    }
    \label{fig:main_analysis}
\end{figure}

Table~\ref{tab:main_results} provides the complete benchmark results.
GPT-6-Astra achieves the strongest overall performance by a large margin,
reaching 80.5\% Choice Avg., compared with 62.8\% for
RynnBrain1.1-122B-A10B, the strongest open-weight model.
Astra leads all seven evaluation dimensions, indicating strong and
relatively balanced performance across recognition, prediction,
alignment, and temporal grounding.
Its strongest result is Frame Matching at 98.3\%, showing that it can
reliably identify the observation corresponding to the queried visual
context.
It also performs strongly on Current Action Recognition (89.4\%) and
Goal-Conditioned Next Action Prediction (83.6\%), suggesting effective
use of both visual evidence and explicit task goals, though Frame
Ordering remains weaker.\par

However, Astra does not saturate the benchmark.
Its performance drops to 76.4\% on Next Action Prediction,
68.3\% on Frame Ordering, 67.5\% on View Matching, and 72.1\% on
Action Time Localization.
These dimensions require reasoning beyond identifying visible content:
the model must infer what happens next, recover temporal order,
associate observations across viewpoints, or localize an action within
the interaction timeline.
Thus, even the strongest model retains substantial room for improvement
on temporal progress understanding in real-world robot manipulation.\par

The remaining models exhibit much more uneven capability profiles.
RynnBrain1.1-122B-A10B, for example, achieves 95.4\% on Frame Matching
but only 32.9\% on Frame Ordering.
This indicates that strong visual correspondence does not necessarily
translate into understanding when events occur or in which order they
unfold.
Doubao-Seed-2.0-Lite shows a similar gap: it performs well on
Current Action Recognition (75.4\%) and Goal-Conditioned Next Action
Prediction (63.1\%), but reaches only 35.9\% on Frame Ordering and
45.1\% on View Matching.
RynnBrain1.1-9B is comparatively strong on future prediction among the
non-Astra models, obtaining 55.0\% on Next Action Prediction, but its
Frame Ordering accuracy remains only 32.8\%.
Together, these results suggest that predicting a plausible next action
and recovering the temporal structure of an observed manipulation remain
distinct challenges. Gemini-3.6-Flash exhibits a notably different profile.
Its recognition performance is weak, with only 24.5\% on Frame Matching
and a Recognition Group score of 29.5\%, yet it achieves the strongest
non-Astra results on View Matching (60.8\%) and Frame Ordering (49.6\%).
This contrast is visible in Figure~\ref{fig:main_analysis}(a,b) and
shows that the benchmark separates recognition-oriented capability from
alignment and temporal order reasoning rather than measuring a single
monolithic notion of visual understanding, which aggregate accuracy
alone would largely conceal.\par

At the lower end of the table, InternVL3-2B, Cosmos3-Edge-2B,
SenseNova-SI-1.1-2B, and Qwen3-VL-2B obtain Choice Avg. scores between
33.2\% and 36.0\%.
Their weaknesses are broad rather than task-specific.
For example, InternVL3-2B and Cosmos3-Edge-2B remain close to the
chance floor on several matching and ordering dimensions, while
SenseNova-SI-1.1-2B achieves an Action Time score of only 0.001.
These models therefore struggle not only with future prediction but
also with basic temporal grounding and cross-observation reasoning. \par

Figure~\ref{fig:main_analysis}(c) further shows that explicit task goals
consistently improve next-action prediction.
Astra improves from 76.4\% to 83.6\%, while all other models also gain
from goal conditioning.
This indicates that knowing the intended task helps constrain plausible
future actions.
Nevertheless, the persistent gap between Next Action and the stronger
recognition tasks shows that goal information alone is insufficient:
models must still connect the current visual state with the temporal
progress of the manipulation. Overall, the benchmark reveals a clear hierarchy of difficulty.
Models are generally strongest at recognizing or matching visible
observations, weaker at predicting future actions, and substantially
less reliable at temporal ordering, cross-view alignment, and precise
action-time grounding.
GPT-6-Astra substantially raises the performance ceiling and shows that
strong performance across all capability groups is achievable, but its
remaining errors on ordering, viewpoint alignment, prediction, and
temporal localization also indicate that real-world manipulation
progress understanding is not yet solved, leaving substantial headroom
for future models.\par

\subsection{Effect of Robot Collection Platform}
\label{sec:platform_comparison}

To separate platform effects from task composition, we compare
GIM and Tianji only on the 12 manipulation tasks recorded on
both platforms.

\input{Tables/compare}

As shown in Table~\ref{tab:platform_comparison} and
Figure~\ref{fig:controlled_analysis}(a), model performance is
broadly consistent across the two robot collection platforms.
Tianji obtains moderately higher scores on all six matched
dimensions, with an overall Choice Avg.\ difference of 5.8 percentage
points, but the magnitude of the shift remains relatively small
compared with the overall variation across models and tasks.
The largest differences occur in Current Action Recognition and
Frame Matching ($\sim$9.6 points), while Frame Ordering differs by
only 2.9 points. Importantly, the overall capability pattern remains similar across
platforms: dimensions that are difficult on GIM remain difficult on
Tianji, and the relative strengths and weaknesses of the evaluated
models are largely preserved.
We therefore view the observed gap as a moderate sensitivity to
collection conditions rather than evidence that benchmark performance
is dominated by a particular robot platform or its specific camera and
gripper hardware.\par

\subsection{Input Ablation}
\label{sec:input_ablation}

\begin{figure}[t]
    \centering
    \includegraphics[width=\linewidth]{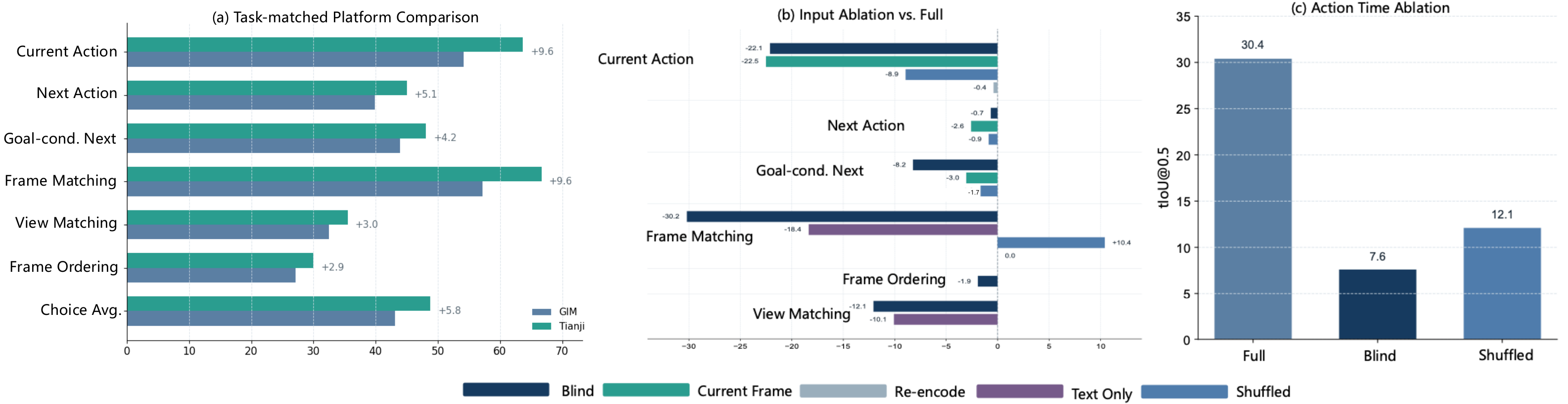}
    \caption{
        Controlled analysis of collection and input effects.
        In (a), gains are Tianji minus GIM in percentage points,
        the opposite sign convention to $\Delta$ in
        Table~\ref{tab:platform_comparison}.
    }
    \label{fig:controlled_analysis}
\end{figure}

We further examine which input cues support benchmark
performance using five open-weight models on a balanced
312-stem subset.
Each ablation is evaluated against the Full condition on matched
questions. Figure~\ref{fig:controlled_analysis}(b) reveals markedly different input dependencies across tasks.
Removing visual observations causes large drops in
Current Action Recognition and Frame Matching
($-22.1$ and $-30.2$ points), whereas Next Action Prediction
changes by less than one point.
Similarly, replacing the observation history with only the
current frame reduces Current Action Recognition by 22.5 points
but has only a small effect on future action prediction.
This suggests that current action understanding relies strongly
on visual history, while next-action prediction can often be
supported by task and action priors even when visual evidence is
entirely unavailable.\par

\input{Tables/abl}

Temporal order produces a complementary effect.
Shuffling frames reduces Current Action Recognition by
8.9 points and Action Time Localization from 30.4 to 12.1
(Figure~\ref{fig:controlled_analysis}(c)),
showing that temporal structure is important for grounding task
progress.
By contrast, Frame Matching increases under shuffling, while
the identity re-encoding control produces almost no change.
The effect therefore reflects task specific use of temporal
order rather than a generic video processing artifact. \par

Overall, the ablations expose a clear distinction between
\emph{recognizing visible evidence} and
\emph{understanding task progression}.
Visual observations are essential for recognition and matching,
temporal context is critical for current state and temporal
grounding, while next-action prediction remains comparatively
insensitive to the exact visual history, relying more on task priors.\par

%% file: Tables/Main_exp.tex
\begin{table}[t]
\centering
\begin{threeparttable}

\caption[Results on \benchmarkname{}.]{
\textbf{Results on \benchmarkname{}.}
Performance across seven evaluation dimensions grouped into recognition,
alignment, and grounding. Group and Choice Avg.\ columns reproduce the aggregates from the
evaluation runs over retained applicable questions. Choice Avg.\ excludes
Action Time, which reports R@1 at tIoU $\geq 0.5$.}

\label{tab:main_results}

\fontsize{8}{9.5}\selectfont
\setlength{\tabcolsep}{2.2pt}

\begin{tabular}{l c cccc c cc c c c}
\toprule

\multirow{2}{*}{Model}
& \multirow{2}{*}{Size}
& \multicolumn{5}{c}{Recognition}
& \multicolumn{3}{c}{Alignment}
& Grounding
& Overall \\

\cmidrule(lr){3-7}
\cmidrule(lr){8-10}
\cmidrule(lr){11-11}
\cmidrule(lr){12-12}

&
& \makecell[c]{Cur.\\Act.}
& \makecell[c]{Next\\Act.}
& \makecell[c]{Next\\+Goal}
& \makecell[c]{Frame\\Match}
& \textit{Group}
& \makecell[c]{View\\Match}
& \makecell[c]{Frame\\Order}
& \textit{Group}
& \makecell[c]{Action\\Time}
& \makecell[c]{Choice\\Avg.} \\

\midrule

\textit{chance floor}
& --
& \textit{0.250}
& \textit{0.250}
& \textit{0.250}
& \textit{0.250}
& \textit{0.250}
& \textit{0.250}
& \textit{0.250}
& \textit{0.250}
& --
& \textit{0.250} \\

\midrule

GPT-6-Astra
& --
& \textbf{0.894}
& \textbf{0.764}
& \textbf{0.836}
& \textbf{0.983}
& \textbf{0.873}
& \textbf{0.675}
& \textbf{0.683}
& \textbf{0.679}
& \textbf{0.721}
& \textbf{0.805} \\

RynnBrain1.1-122B-A10B
& 122B
& 0.752
& 0.511
& 0.588
& 0.954
& 0.708
& 0.536
& 0.329
& 0.442
& 0.512
& 0.628 \\

Doubao-Seed-2.0-Lite
& --
& 0.754
& 0.538
& 0.631
& 0.921
& 0.717
& 0.451
& 0.359
& 0.409
& 0.577
& 0.624 \\

RynnBrain1.1-9B
& 9B
& 0.730
& 0.550
& 0.599
& 0.849
& 0.687
& 0.444
& 0.328
& 0.392
& 0.405
& 0.598 \\

Qwen3-VL-235B-A22B
& --
& 0.697
& 0.471
& 0.561
& 0.797
& 0.637
& 0.392
& 0.304
& 0.352
& 0.445
& 0.551 \\

Qwen3-VL-32B
& 32B
& 0.709
& 0.509
& 0.566
& 0.720
& 0.630
& 0.307
& 0.329
& 0.317
& 0.469
& 0.536 \\

Qwen3-VL-30B-A3B
& 30B-A3B
& 0.684
& 0.426
& 0.495
& 0.853
& 0.621
& 0.313
& 0.290
& 0.302
& 0.418
& 0.525 \\

RynnBrain1.1-2B
& 2B
& 0.618
& 0.390
& 0.464
& 0.848
& 0.587
& 0.413
& 0.264
& 0.346
& 0.151
& 0.514 \\

Qwen3-VL-8B
& 8B
& 0.653
& 0.459
& 0.501
& 0.690
& 0.580
& 0.324
& 0.284
& 0.306
& 0.385
& 0.498 \\

Qwen3-VL-8B-Thinking
& 8B
& 0.608
& 0.453
& 0.508
& 0.715
& 0.575
& 0.315
& 0.277
& 0.298
& 0.335
& 0.492 \\

MiMo-VL-7B-RL
& 7B
& 0.570
& 0.428
& 0.464
& 0.778
& 0.565
& 0.283
& 0.294
& 0.288
& 0.103
& 0.482 \\

RynnBrain-2B
& 2B
& 0.574
& 0.377
& 0.438
& 0.656
& 0.516
& 0.344
& 0.254
& 0.303
& 0.188
& 0.452 \\

Cosmos-Reason2-2B
& 2B
& 0.578
& 0.435
& 0.458
& 0.465
& 0.486
& 0.304
& 0.271
& 0.289
& 0.383
& 0.427 \\

Gemini-3.6-Flash
& --
& 0.284
& 0.308
& 0.347
& 0.245
& 0.295
& 0.608
& 0.496
& 0.558
& 0.187
& 0.374 \\

Qwen3-VL-2B
& 2B
& 0.509
& 0.323
& 0.335
& 0.396
& 0.394
& 0.289
& 0.270
& 0.281
& 0.223
& 0.360 \\

SenseNova-SI-1.1-2B
& 2B
& 0.409
& 0.349
& 0.415
& 0.398
& 0.393
& 0.266
& 0.286
& 0.275
& 0.001
& 0.358 \\

Cosmos3-Edge-2B
& 2B
& 0.503
& 0.401
& 0.421
& 0.237
& 0.390
& 0.222
& 0.204
& 0.214
& 0.253
& 0.339 \\

InternVL3-2B
& 2B
& 0.433
& 0.340
& 0.423
& 0.260
& 0.364
& 0.258
& 0.262
& 0.260
& 0.023
& 0.332 \\

\bottomrule
\end{tabular}
\begin{tablenotes}[flushleft]
\footnotesize
\item The multiple-choice chance floor is 0.25. A comparable R@1 baseline
for Action Time is not reported; the mean-tIoU baseline is omitted because
it uses a different metric.
\end{tablenotes}
\end{threeparttable}
\end{table}

%% file: Tables/compare.tex
\begin{table}[t]
\centering
\small
\setlength{\tabcolsep}{4.2pt}
\begin{threeparttable}

\caption[Task-matched comparison across robot collection platforms.]{
\textbf{Task-matched comparison across robot collection platforms.}
We restrict GIM and Tianji gripper recordings to the 12 manipulation
tasks available on both platforms and report question-weighted accuracy
to isolate platform-specific effects. Scores are percentages;
$\Delta = \mathrm{GIM} - \mathrm{Tianji}$ in percentage points.
Platform accuracy is pooled over the 17 models with per-question records;
the final column counts platform wins across all 18 models.
}
\label{tab:platform_comparison}

\begin{tabular}{lcccc}
\toprule
\textbf{Evaluation dimension}
& \textbf{GIM}
& \textbf{Tianji}
& $\boldsymbol{\Delta}$
& \textbf{Tianji $>$ GIM (models)} \\
\midrule

Current Action
& 54.1
& 63.7
& -9.6
& 17 / 18 \\

Next Action
& 39.9
& 45.0
& -5.1
& 14 / 18 \\

Goal-conditioned Next Action
& 43.9
& 48.1
& -4.2
& 14 / 18 \\

Frame Matching
& 57.2
& 66.7
& -9.6
& 17 / 18 \\

View Matching
& 32.5
& 35.5
& -3.0
& 15 / 18 \\

Frame Ordering
& 27.1
& 30.0
& -2.9
& 17 / 18 \\

\midrule

\textbf{Choice Avg.}
& \textbf{43.1}
& \textbf{48.8}
& \textbf{-5.8}
& \textbf{18 / 18} \\

\bottomrule
\end{tabular}

\end{threeparttable}
\end{table}

%% file: Tables/abl.tex
\begin{table}[!t]
\centering
\small
\setlength{\tabcolsep}{3.0pt}

\caption{
Input ablation on \benchmarkname
}
\label{tab:input_ablation}

\resizebox{\linewidth}{!}{%
\begin{tabular}{lccccccc}
\toprule
\textbf{Input}
& \makecell[c]{\textbf{Current}\\\textbf{Action}}
& \makecell[c]{\textbf{Next}\\\textbf{Action}}
& \makecell[c]{\textbf{Goal-cond.}\\\textbf{Next}}
& \makecell[c]{\textbf{Frame}\\\textbf{Match}}
& \makecell[c]{\textbf{Frame}\\\textbf{Order}}
& \makecell[c]{\textbf{View}\\\textbf{Match}}
& \makecell[c]{\textbf{Action}\\\textbf{Time}} \\
\midrule

\textbf{Full}
& \textbf{62.6}
& \textbf{40.0}
& \textbf{49.3}
& \textbf{46.0}
& \textbf{29.4}
& \textbf{34.9}
& \textbf{30.4} \\

\midrule

Blind
& 40.5 {\scriptsize(-22.1)}
& 39.3 {\scriptsize(-0.7)}
& 41.1 {\scriptsize(-8.2)}
& 15.8 {\scriptsize(-30.2)}
& 27.5 {\scriptsize(-1.9)}
& 22.8 {\scriptsize(-12.1)}
& 7.6 {\scriptsize(-22.8)} \\

Text Only
& --
& --
& --
& 27.6 {\scriptsize(-18.4)}
& --
& 24.8 {\scriptsize(-10.1)}
& -- \\

Current Frame Only
& 40.0 {\scriptsize(-22.5)}
& 37.5 {\scriptsize(-2.6)}
& 46.2 {\scriptsize(-3.0)}
& --
& --
& --
& -- \\

Shuffled Frames
& 53.6 {\scriptsize(-8.9)}
& 39.2 {\scriptsize(-0.9)}
& 47.6 {\scriptsize(-1.7)}
& 56.4 {\scriptsize(+10.4)}
& --
& --
& 12.1 {\scriptsize(-18.3)} \\

Identity Re-encode
& 62.1 {\scriptsize(-0.4)}
& --
& --
& 45.9 {\scriptsize(+0.0)}
& --
& --
& -- \\

\bottomrule
\end{tabular}%
}
\end{table}

%% file: Sections/5_Conclusion.tex
\section{Conclusion}
\label{sec:conclusion}

We introduced \benchmarkname{}, a real-robot benchmark for evaluating temporal task understanding from progressively revealed visual observations, covering 39 manipulation scenarios and seven complementary evaluation tasks. Our results reveal a persistent gap between recognizing visible content and understanding how a physical task unfolds over time. Even strong multimodal models exhibit substantial weaknesses in temporal ordering, future action prediction, viewpoint alignment, and temporal localization, while controlled ablations further reveal distinct dependencies on observation history, temporal order, and task priors. These findings highlight the importance of evaluating temporal capabilities individually rather than relying solely on aggregate performance. We hope \benchmarkname{} provides a useful testbed for developing multimodal systems that can more reliably reason about task progress during ongoing real-world robot interaction.

%% file: Sections/6_appendix.tex
\begin{figure}[H]
    \centering
    \includegraphics[width=\linewidth]{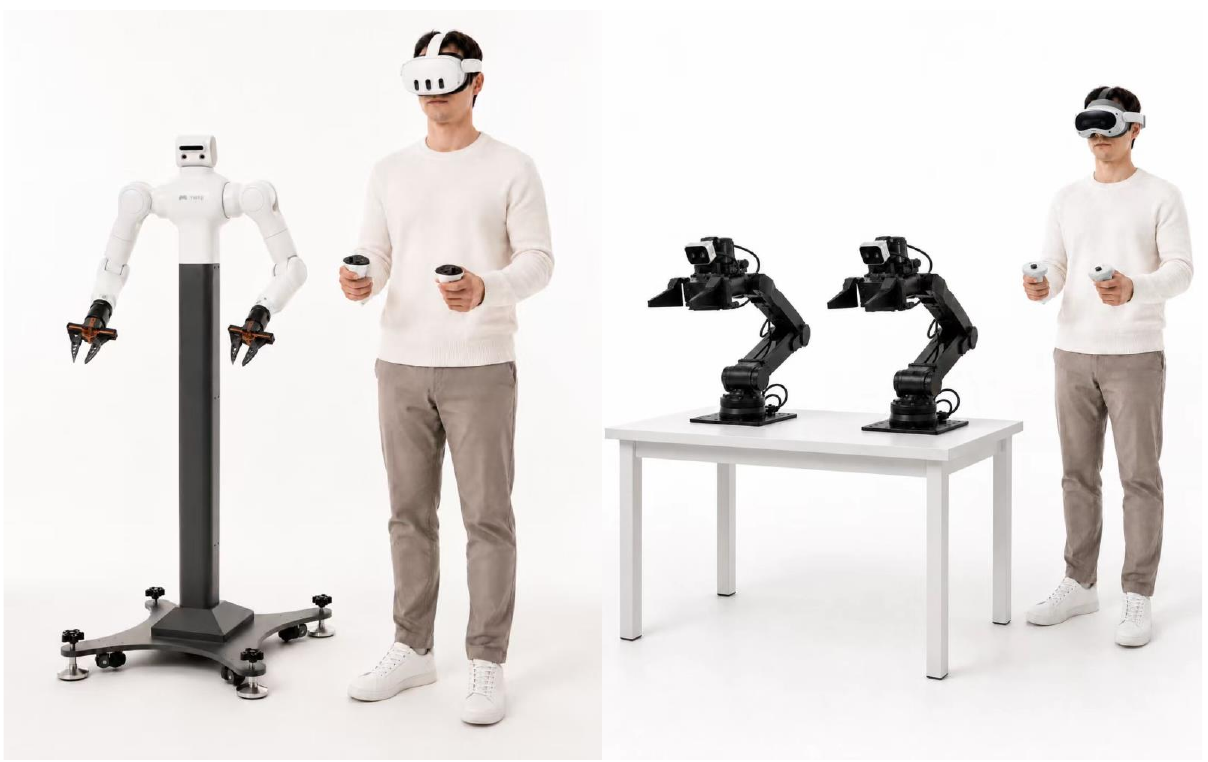}
    \caption{
    \textbf{Hardware platforms used in our benchmark.}
    We employ two real-world robotic manipulation setups with different
    embodiments and workspace configurations. Both platforms support
    VR-based human teleoperation for collecting manipulation trajectories.
    }
    \label{fig:hardware_platforms}
\end{figure}

\label{app:conditional}
\paragraph{Hardware Platforms.}
Our benchmark is built upon two real-world robotic manipulation platforms,
covering different robot embodiments and workspace configurations.
As illustrated in Fig.~\ref{fig:hardware_platforms}, the first platform is
a dual-arm robotic system for whole-workspace manipulation, while the second
platform consists of two tabletop robotic manipulators.
Both platforms are operated through a VR-based teleoperation interface,
where human operators use a head-mounted display and handheld controllers
to perform manipulation tasks.
These complementary hardware setups provide diverse real-world robot
trajectories for benchmark construction and evaluation.

\paragraph{Conditional Action Time Evaluation.}
The Action Time task requires models to answer with temporal intervals in
seconds. Although the corrected prompt states the total duration of the queried
video, some systems still return intervals as fractions of the video duration.
This behavior is a unit error rather than a temporal localization error. We
therefore report a conditional Action Time score that evaluates only the
answers that are given in seconds.

Let $\mathcal{S}$ denote the set of localization queries for which the model
returns a parseable interval in seconds. The conditional score is computed as
\[
    \frac{1}{|\mathcal{S}|}
    \sum_{i \in \mathcal{S}}
    \mathbf{1}\left[
        \mathrm{tIoU}(\hat{I}_i, I_i^*) \geq 0.5
    \right].
\]
The reference interval and scoring threshold are unchanged from the main
evaluation. The only difference is that normalized interval answers are
excluded from the denominator.

\clearpage
\paragraph{Qualitative Action Time Examples.}
Figure~\ref{fig:action_time_examples} presents six qualitative examples
of Action Time localization from the GIM and Tianji platforms,
comparing the predictions of five models with the ground-truth intervals.
The examples illustrate both accurate temporal alignment and different
failure cases, including overly broad intervals, predictions that do
not overlap the reference interval, and timestamps beyond the video
duration.

\begin{figure}[H]
    \centering
    \includegraphics[width=0.97\linewidth]{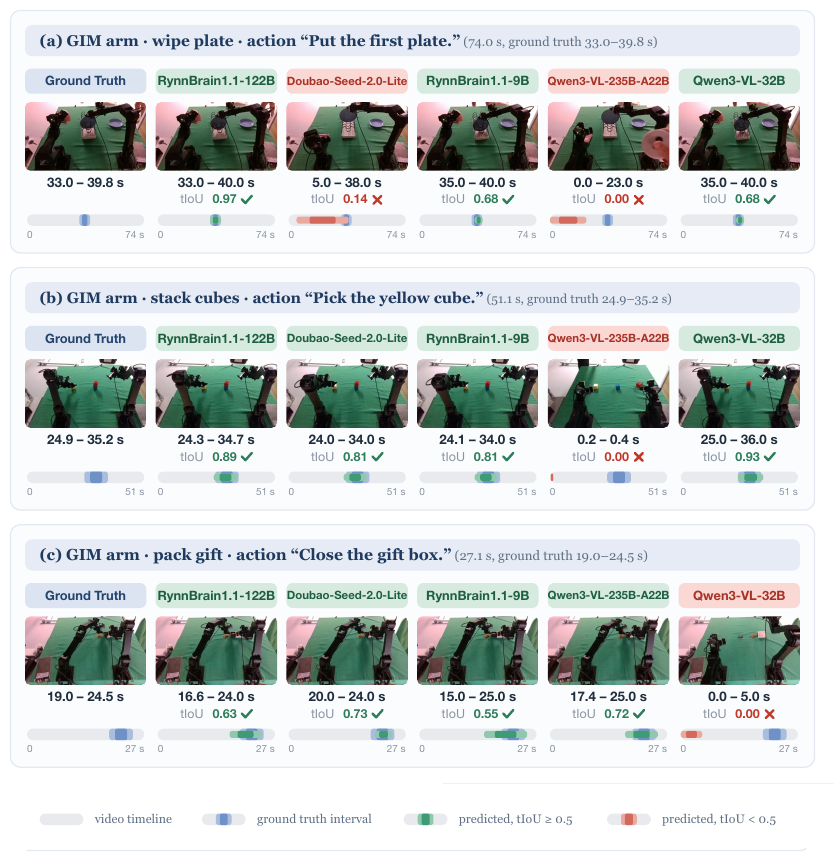}
    \caption{\textbf{Qualitative examples of Action Time localization.}
    Panels (a)--(c) show the GIM examples at an enlarged scale.
    Ground-truth and predicted intervals are unchanged from the original examples.}
    \label{fig:action_time_examples}
\end{figure}
\clearpage
\begin{figure}[t]
    \ContinuedFloat
    \centering
    \includegraphics[width=0.97\linewidth]{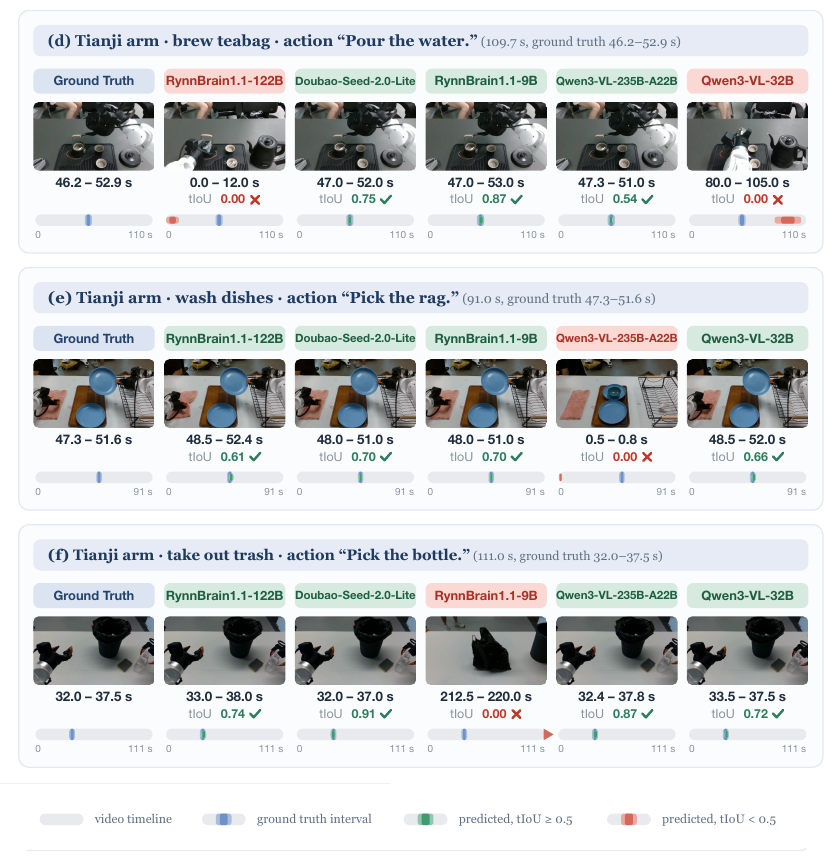}
    \caption{\textbf{Qualitative examples of Action Time localization (continued).}
    Panels (d)--(f) show the Tianji examples at an enlarged scale.}
\end{figure}
\clearpage

\paragraph{RoboChrono Evaluation Pipeline.}
  Algorithm~\ref{alg:robochrono} summarizes the complete RoboChrono evaluation
  pipeline across different models, scenarios, and evaluation dimensions.
  For each question, the pipeline prepares the visual input, performs model
  inference, parses either a multiple-choice answer or a temporal interval,
  and computes the corresponding accuracy or $\mathrm{tIoU}@0.5$ score.
  Finally, all question-level results are aggregated into the evaluation report.

\begin{algorithm}[H]
  \caption{RoboChrono Evaluation}
  \label{alg:robochrono}
  \KwIn{Models $\mathcal{M}$, scenarios $\mathcal{S}$, dimensions $\mathcal{D}$}
  \KwOut{Evaluation report $\mathcal{R}$}

  Validate dataset and model configurations\;
  Initialize result set $\mathcal{R}\leftarrow\emptyset$\;

  \ForEach{model $m\in\mathcal{M}$}{
      Initialize model adapter $A_m$\;

      \ForEach{scenario $s\in\mathcal{S}$}{
          \ForEach{dimension $d\in\mathcal{D}$}{
              Load questions $Q_{s,d}$\;

              \ForEach{question $q\in Q_{s,d}$}{
                  \If{$q$ has already been completed}{
                      continue\;
                  }

                  $x\leftarrow\mathrm{PrepareInput}(q)$\;
                  $y\leftarrow A_m.\mathrm{Generate}(x)$\;

                  \eIf{$d=\texttt{action\_time}$}{
                      $\hat{t}\leftarrow\mathrm{ParseInterval}(y)$\;
                      $z\leftarrow
                      \mathbb{I}\!\left(
                      \mathrm{tIoU}(\hat{t},t_q)\geq0.5
                      \right)$\;
                  }{
                      $\hat{c}\leftarrow\mathrm{ParseChoice}(y)$\;
                      $z\leftarrow\mathbb{I}(\hat{c}=c_q)$\;
                  }

                  Save $(m,s,d,q,y,z)$\;
              }

              $\mathrm{Score}_{m,s,d}
              \leftarrow
              \frac{1}{|Q_{s,d}|}
              \sum_{q\in Q_{s,d}}z_q$\;
          }
      }
  }

  Aggregate all scores into $\mathcal{R}$\;
  \KwRet $\mathcal{R}$\;
  \end{algorithm}